 \documentclass[pmlr,twocolumn,10pt]{jmlr} 

\usepackage{amsmath}
\DeclareMathOperator*{\argmax}{arg\,max}

\usepackage{booktabs}
\usepackage{multirow}

\usepackage{siunitx}

\theorembodyfont{\upshape}
\theoremheaderfont{\scshape}
\theorempostheader{:}
\theoremsep{\newline}

\jmlrvolume{LEAVE UNSET}
\jmlryear{2023}
\jmlrsubmitted{LEAVE UNSET}
\jmlrpublished{LEAVE UNSET}
\jmlrworkshop{Machine Learning for Health (ML4H) 2023} 

\title[Mallows model for characterising symptom sequences]{Bayesian inference of a new Mallows model for characterising symptom sequences applied in primary progressive aphasia}

\author{
\Name{Beatrice Taylor} \Email{beatrice.taylor.20@ucl.ac.uk}\\
\addr UCL Centre for Medical Image Computing, Department of Computer Science, University College London, UK
\AND
\Name{Cameron Shand} \Email{}\\
\addr UCL Centre for Medical Image Computing, Department of Computer Science, University College London, UK
\AND
\Name{Chris J. D. Hardy} \Email{}\\
\addr UCL Queen Square Institute of Neurology Dementia Research Centre, University College London, UK
\AND
\Name{Neil P.~Oxtoby} \Email{}\\
\addr UCL Centre for Medical Image Computing, Department of Computer Science, University College London, UK
}

\begin{document}

\maketitle

\begin{abstract}
Machine learning models offer the potential to understand diverse datasets in a data-driven way, powering insights into individual disease experiences and ensuring equitable healthcare. In this study, we explore Bayesian inference for characterising symptom sequences, and the associated modelling challenges. We adapted the Mallows model to account for partial rankings and right-censored data, employing custom MCMC fitting. Our evaluation, encompassing synthetic data and a primary progressive aphasia dataset, highlights the model's efficacy in revealing mean orderings and estimating ranking variance. This holds the potential to enhance clinical comprehension of symptom occurrence. However, our work encounters limitations concerning model scalability and small dataset sizes. 
\end{abstract}

\section{Introduction}
\label{sec:intro}
An important aspect of data-driven healthcare is making sense of which symptoms will occur when. With the global ageing population comes an increased prevalence of age-related chronic conditions where symptoms tend to accumulate over time --- the number of people living with dementia globally is expected to exceed 150 million by 2050 \citep{Nichols2022Estimation2019}. Patients' symptoms can be highly heterogeneous, which is where advanced modelling and machine learning methods can be used to improve health outcomes. Heterogeneous presentation is particularly pronounced in the rarer dementias which are associated with unusual symptoms \citep{Marshall2018PrimaryApproach}, and a higher caregiver burden \citep{Brotherhood2020ProtocolImpact}. Precise prediction of symptom occurrences post-diagnosis could alleviate care responsibilities and enhance our ability to tailor support to patients.

Existing disease progression models for neurodegeneration have typically been used with imaging data \citep{Oxtoby2017ImagingDisease}, with proven application in identifying data-driven trajectories of brain volume change \citep{Young2018UncoveringInference}. We were interested if similar probabilistic models could be applied to healthcare measures in order to understand changes in clinical presentation \citep{Beckett1993MaximumData}. 

This work builds on the research by \cite{Huang2012ProbabilisticDisease} and \cite{Young2015MultipleProgression} that combined disease progression models with the Mallows model for ranking disease events. Employing a Bayesian framework, we infer the parameters of a Mallows distribution \citep{mallows1957non}, which is analogous to a Gaussian distribution for rankings. Model fitting employed a Markov Chain Monte Carlo (MCMC) approach.

The primary contributions of our research are as follows:
\begin{itemize}
    \item \textit{Enhancement of an existing model} --- adapted Mallows model to effectively handle censored data and partial rankings.
    \item \textit{Application to a novel healthcare dataset} --- applied our model to a previously unexplored dataset consisting of symptom questionnaires. 
\end{itemize}

To evaluate the performance of our model, we first conducted tests on synthetic data with added noise. Subsequently, we assessed the model's effectiveness using a real-world healthcare questionnaire dataset \citep{Hardy2023Symptom-ledAphasia} from patients diagnosed with primary progressive aphasia (PPA), a rare language led dementia associated with atypical symptoms \citep{Gorno-Tempini2011ClassificationVariants}.

\section{Method}

\subsection{Mathematical model}
From a set of disease-related symptoms, each patient will typically present with a subset, with multiple symptoms occurring concurrently. These individual symptom profiles can be viewed as variations of a central ranking at a group level. In our model, we understand these patient-specific symptom profiles as partial rankings that follow a Mallows distribution.

Consider we have $M$ survey responses. As part of the survey each respondent $m \in M$ was asked to rank $n$ symptoms, $S=\{S_1,S_2,...,S_n\}$ in $l$ positions, $\{1, 2, ...,l\}$ according to if and when the symptoms occurred. 

Individuals might not rank every symptom either due to randomness or due to having only experienced a subset of symptoms at the time of answering the survey. To account for this we define the subset $\hat{S}_m \subseteq S$, with $r_m=|\hat{S}_m|$, of symptoms for which individual $m$'s response was recorded. Each questionnaire response is a partial ranking of events represented by the mapping:
\begin{align}
    \sigma_m:\{\hat{S}_{m,1}, ..., \hat{S}_{m,r}\} \longmapsto \{1, ..., l\}^{r_m}.
\end{align}
For a participant $m$, their partial ranking of events is given by: 
\begin{align}
X_m &=\{ \{\sigma^{-1}_m(1) \} ,..., \{ \sigma^{-1}_m(l)\} \},
\end{align}
where $\sigma^{-1}_m(l)$ refers to the set of events assigned to stage $l$.
The Mallows model is a probability distribution for rankings parameterised by a central ranking $\pi_0$ and a spread parameter $\lambda$ \citep{Fligner1986DistanceModels, Tang2019MallowsRegeneration}. The probability density function is given by:
\begin{align}\label{eq:pdf_PM}
f_{\pi_0, \lambda}(x) &= \psi(\lambda) e^{- \frac{1}{\lambda} d(x, \pi_0)}, \\
\psi (\lambda) &= \sum_{\pi \in \mathfrak{S_n}}e^{-\frac{1}{\lambda} d(\pi, \pi_0}),
\end{align}

where $\psi(\lambda)$ is a normalising function summed over the set of all possible rankings $\mathfrak{S_n}$, and  $d(\pi, \pi_0)$ is a distance metric generally taken to be Kendall's Tau \citep{Tang2019MallowsRegeneration}. For $\lambda \geq 0$, the central ranking $\pi_0$ is the mode, and as $\lambda \longrightarrow 0$ the model is concentrated at $\pi_0$. When $\lambda=0$ it is the uniform distribution \citep{Fligner1986DistanceModels}. We describe a set of ranks as having a strong consensus for $\lambda \leq \varepsilon$ and weak consensus as $\lambda \longrightarrow \infty$ \citep{Ali2012ExperimentsWhen}. 

We adapt the model to account for partial rankings by using the Kendall's Tau distance metric with penalty parameter $p$ \citep{Fagin2003ComparingLists}: 
\begin{equation}\label{eq:partial_tau}
d_p(\pi, \pi_0) = |\beta_D|+p*|\beta_E|,
\end{equation}
where $\beta_D$ is the set of discordant pairs, and $\beta_E$ is the set of pairs that have equal position in one ranking but not in the other. The choice of $p \in [0,1]$ determines the weighting of partial ranks --- for simplicity we fix  $p=0.5$ \citep{Cohen1997LearningThings}, further details in Appendix \ref{apd:models}. To account for comparison of rankings where items are unranked due to censoring, or missingness, we drop the comparison of pairs from the calculation (where one or more rank is missing). 

It is worth noting the difference in the size of the space between a traditional Mallow's model and one that is partially ranked. For a fully ranked model where rankings are permutations of the range of numbers up to the maximum rank $n$, the space of rankings is: $|\mathfrak{S_n}|=n!$. However in the case of partial rankings the space of possible rankings is significantly larger with: 
\begin{align}\label{eq:cardinality}
    |\mathfrak{S_n}|=l^n. 
\end{align}

The likelihood of a patients data $X_m$ given $\pi_0, \lambda$ is:
\begin{equation}\label{eq:likelihood1}
p(X_m |\pi_0, \lambda) = f_{\pi_0, \lambda}(X_m).    
\end{equation}
We assume that data from patients is independent, obtaining the likelihood for dataset $X$ as: 
\begin{equation}\label{eq:likelihood2}
p(X |\pi_0, \lambda) = \prod_m f_{\pi_0, \lambda}(X_m)    
\end{equation}
According to Bayes theorem, the model posterior is given by: 
\begin{equation}
    p(\pi_0, \lambda | X) \propto p(\pi_0, \lambda) p(X |\pi_0, \lambda)
\end{equation}
with joint prior,
\begin{equation}
    p(\pi_0, \lambda) = p(\pi_0 | \lambda)*p(\lambda).
\end{equation}
The prior distributions on $\pi_0$ and $\lambda$ are taken to be:
\begin{align}
\lambda &\sim truncatednorm(0,1),\\
\pi_0 &\sim mallows(\pi_{\text{init}}, \lambda).
\end{align}
with $\pi_{\text{init}}$ informed by clinical input. 
We justify a choice of an informative prior based on the large distribution space. 

\subsection{Model fitting}
We use an MCMC algorithm to sample from $p(\pi_0, \lambda | X)$. Details are in Appendix \ref{apd:models}. We derive a maximally likely ranking of symptoms $\pi_0$, and the corresponding spread parameter $\lambda$ using the MAP estimate of a set of $1,000$ MCMC samples. 

The MAP estimates are defined as: 
\begin{align}
\pi_{0, \text{MAP}} &= \argmax_{\pi_0} P(D|\pi_0, \lambda) P(\pi_0, \lambda)  \\
\lambda_{0, \text{MAP}} &= \argmax_{\lambda} P(D|\pi_0, \lambda) P(\pi_0, \lambda).
\end{align}

\section{Experimental set-up}

We performed two experiments. First, we perform parameter estimation on synthetic data that we generate to mimic our real-world data of interest, providing a ground truth to assess our proposed method's accuracy. Second, we estimate model parameters for real-world healthcare data from a study of people living with PPA \citep{Hardy2023Symptom-ledAphasia}. 

\subsection{Synthetic data}
For a given central ordering, $\pi_0$, with length $n$, and maximum rank of $l$ and spread parameter, $\lambda$, we generated synthetic datasets of size $M$ by sampling from the space of possible rankings $\mathfrak{S_n}$ according to equation \ref{eq:pdf_PM}. We simulated missing data due to the right censoring issue described earlier (details in Appendix \ref{apd:models}). Figure \ref{fig:spread_param} shows the distribution of simulated data about the central ordering as a function of the Mallows spread parameter, $\lambda$.

\begin{figure}[h]
\floatconts
  {fig:spead_param}
  {\caption{Example of how the synthetic Mallows-distributed data varies as a function of the the spread parameter $\lambda$.} \label{fig:spread_param}}
  {\includegraphics[width=0.9\linewidth]{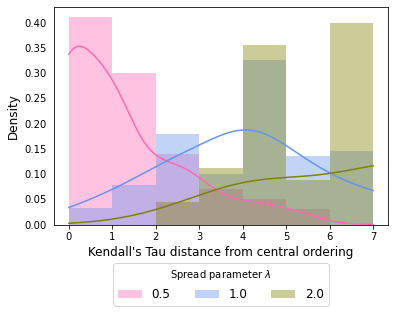}}
\end{figure}


We used the model to infer the parameters of the synthetic data and compared the result to the true parameters used to generate the data. Table \ref{table:synthetic_data_results} demonstrates the models utility in a dataset of size $M=100$ with rankings of length $n=8$, and maximum rank of $l=4$. The ranking length and maximum rank were chosen based on the healthcare data in Section \ref{sec:healthcare}. The values are averaged over $12$ repeats with random initialisation.

\subsection{Healthcare data}\label{sec:healthcare}

Data was collected from both carers to people living with svPPA, and people with svPPA in the UK and Australia. Further details of the original clinical study can be found in \cite{Hardy2023Symptom-ledAphasia}. The full questionnaire included $n=72$ symptoms that individuals were asked to rank. Based on the results in the synthetic dataset we realised it was unrealistic to model the full dimensionality of the data, as such we chose a subset of symptoms relating to personal care and well being, and which at least 30\% of the cohort had ranked. The resultant list of eight symptoms is given in Appendix \ref{apd:results}. In total there were $m=30$ individuals in the dataset.

For model comparison we analysed this dataset using the ordinal subtype and stage inference (SuStaIn) algorithm \citep{young2021ordinal} with a single subtype, equivalent to an ordinal version of the Event Based Model (EBM) \citep{fonteijn2012event}. This model is based on the assumption that disease events that have been experienced by more individuals in the dataset occur earlier in the disease progression. It uses a Bayesian model and MCMC to calculate the MAP sequence of disease events. To accommodate SuStaIn, the data is simplified to be either 1 if the event was ranked by the respondent, or 0 otherwise. The model assumes that there are as many stages of disease as events, and hence ranks the events occurring as per a trajectory, or permutation. 

\section{Results}

\subsection{Synthetic data}

\begin{table}[h]
    \centering
    \begin{tabular}{p{0.3cm}p{1.6cm}p{1.3cm}p{1.9cm}}
        $\lambda_0$ & \% missing & MAE $\lambda_0$ & $\overline{d_p(\pi_0, \hat{\pi}_{0})}$ \\
        \midrule\midrule
        0.5 & 0   & 0.024 & 0.167 \\
            \midrule
        1.0 & 0   & 0.049 & 0.167 \\
            & 10 & 0.017 & 0 \\ \midrule 
        2.0 & 0   & 0.176   & 0.250 \\
            & 10 & 0.049      & 0.5 \\
        \bottomrule
    \end{tabular}
\caption{Results from synthetic data experiments. Mean Absolute Error (MAE) in Mallows model spread $\lambda_0$ and mean distance from central ranking $\pi_0$ increased with missing data \% ($\hat{\pi}_0$ denotes the MAP estimate). The central ordering was $\pi_0=[1, 2, 2, 3, 3, 3, 3, 4 ]$.}
\label{table:synthetic_data_results}
\end{table}

Our model is able to accurately infer the model parameters. As expected, model results worsen as a function of the percent of missing data, and the spread parameter $\lambda_0$. The mean absolute error of zero, for $\lambda_0 = 1.0$ and 10\% missing, is out of pattern with the rest of the results, but arises from the model finding the exact central ordering in each experiment repetition. 

Even in this small synthetic data example, the size of the sample space is $8^4=4096$. Due to the exponential nature of the sample space (\ref{eq:cardinality}) the number of calculations required scales poorly, especially since the model requires a repeat calculation of the normalising constant at each model iteration, with each calculation of the normalising constant being a sum over $l^n$ values. 

\subsection{Healthcare data}
\begin{figure}[htbp]
\floatconts
  {fig:nodes}
  {\caption{Data-driven symptom staging for the set of 30 individuals with svPPA using the ordinal EBM.} \label{fig:PPA_EBM}}
  {\includegraphics[width=0.95\linewidth]{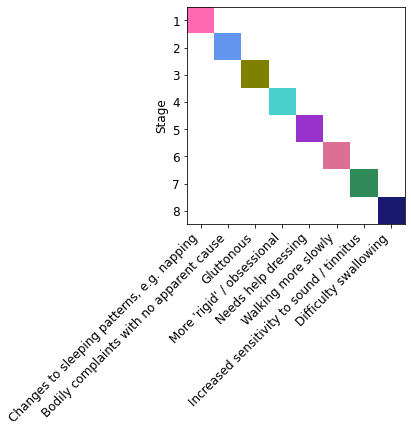}}
\end{figure}

Figure \ref{fig:PPA_EBM} shows the results of modelling the PPA dataset using the baseline model - the ordinal EBM. Each symptom in this model is assigned a unique stage, resulting in a permutation ranking. 

\begin{figure}[htbp]
\floatconts
  {fig:svPPA_mallows}
  {\caption{Data-driven symptom staging for the set of 30 individuals with svPPA, estimated by our partial Mallows model.} \label{fig:PPA_stages}}
  {\includegraphics[width=0.95\linewidth]{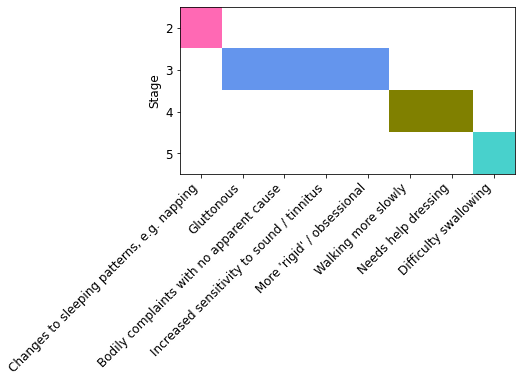}}
\end{figure}

Figure \ref{fig:PPA_stages} shows the resultant central ordering for the dataset using our partially-ordered Mallows model. The prior central ordering used to initialise the model from was based on guidance from PPA researchers. We initialised the spread parameter from $\lambda_{init}=1$. 

The ranking of symptoms shown in Figure \ref{fig:PPA_stages} was the same as the clinically informed ranking of symptoms suggested by the PPA researchers. Compared to Figure \ref{fig:PPA_EBM}, in Figure \ref{fig:PPA_stages} we see the visual grouping of symptoms which co-occur. Both models identify 'changes to sleeping patterns' as the first symptom, and 'difficulty swallowing' as the final symptom. The stage numbering starts from 2 in Figure \ref{fig:PPA_stages}, as the stage numbering is a direct reflection of how the symptoms ranks were assigned in the questionnaire responses. The numbering in Figure \ref{fig:PPA_EBM} is abstracted from this, as the number of stages relates to the total number of symptoms ranked. 

\section{Discussion}
We adapted the Mallows model for partially ranked and censored data, and applied a Bayesian method to estimate model parameters. We used the method to identify the central ranking of symptoms in a novel healthcare dataset collected in PPA. 

Finding the optimal ordering proves to be computationally challenging, being NP-hard even in cases with just four votes within a fully-ordered model \citep{Ali2012ExperimentsWhen, Cohen1997LearningThings}. This computational complexity contributes to scalability issues within this modeling framework. The current bottleneck arises from calculating the normalising constant across the entire distribution space. While a more accurate analytic approximation exists for the fully ranked Mallows model \citep{Fligner1986DistanceModels}, a counterpart for the partially ranked model does not yet exist to the best of our knowledge.

There was perhaps an imbalance between model complexity and data size --- especially for this rare disease. However, the partial rankings model and our new method for handling missing data are exciting prospects for future work in other areas. We explored tolerance of our method to missing (synthetic) data, but further detailed experiments are warranted. In particular rigorous ablation studies, to test model performance as a function of censoring, and evaluation using the widely applicable information criterion (WAIC) \citep{Vehtari2015PracticalWAIC}. Specific to our real-world experiments, the questionnaire's phrasing itself introduced a bias in data acquisition --- symptoms were ordered in a sequence arising from clinical experience. However, this \textit{could} be seen as a kind of implicit prior rather than a bias, but we acknowledge the potential influence on survey respondents' perceived temporal relationships among the symptoms.

Considering real-world healthcare datasets, it is reasonable to anticipate modest consensus levels equating to larger values of the spread parameter $\lambda$. Data with pronounced consensus would permit more efficient central ordering estimation, potentially through alternative Kemeny optimisation methods \citep{Ali2012ExperimentsWhen}. However, in scenarios marked by weak consensus, the necessity of substantial datasets becomes evident. Future research should consider alternative datasets, such as the activities of daily living questionnaires (ADLQ). The ADLQ  inherently establishes latent event rankings, and is commonly used in clinics, thus offering access to larger datasets. Additionally, the potential for application in wearable technology, used for tracking daily activities in dementia patients and offering objective data \citep{Ray2019APatients}, merits exploration.

\subsection{Conclusion}
In conclusion, our study offers insight into the use of Bayesian inference for uncovering symptom sequences. In particular we developed a new Mallows model which allowed us to model partially ranked and censored data. This is an NP-hard problem and our modest results reflect this --- but we are optimistic that solving the optimisation problem will improve performance. Furthermore, we think the idea of using statistical analyses to understand the lived experience of disease warrants further exploration, and we imagine this could have applications in data from wearable technology.


\newpage
\bibliography{references_mallows}

\newpage

\appendix

\section{Model details}\label{apd:models}

\subsection{Model parameters}


The introduction of the hyperparameter $p$ allows us to weight the contribution of partial rankings with the Kendall's Tau measure. For $p \in [0,0.5)$ Kendall’s Tau distance metric with penalty parameter p is a `half metric', failing to satisfy the triangle inequality. For $p \in [0.5,1]$ it is a full metric \citep{Fagin2003ComparingLists}. As per \cite{Fagin2003ComparingLists} the choice of $p=0.5$ can be thought of as a 'neutral penalty score' for partial ranks, indicating that whilst not of equal importance as discordant pairs (i.e. $p=1$), partial ranks should be considered distinct from concordant pairs (i.e. $p=0$). A similar weighting is also utilised in \cite{Cohen1997LearningThings}. Furthermore, a choice of $p \in [0.5, 1]$ is desirable as it avoids modelling complications arising from using a half metric. 

\subsection{MCMC algorithm}

The MCMC algorithm utilises Gibb's sampling similar to the method used by \cite{Young2015MultipleProgression}. The MCMC algorithm proceeds as given in Algorithm \ref{algo:bayesianMallows}. 


\begin{algorithm}
\floatconts
    {alg:MCMC_MAP}
    {\caption{MCMC MAP optimisation}}
    {
\begin{enumerate*}
    \item $\pi_0 \gets \pi_{\text{init}}$
    \item $\lambda_0 \gets \lambda_{init}$
    \item $p_{\pi_{0}} \gets P(\pi_0, \lambda_0|X)$ 
    \item $p_{\lambda_{0}} \gets P(\pi_0, \lambda_0|X)$ 
    \item \For{$t=1$ to $R$}{Sample $\pi' \sim mallows(\pi_{t-1}, \lambda_{t-1})$\\
    $p_{\pi'} \gets P(\pi', \lambda_{t-1}|X)$ \\
    $\pi_t \gets \pi'$ with probability $\alpha=min(1,\frac{p_{\pi_{t-1}}}{p_{\pi'}})$\\
    Sample $\lambda' \sim truncatednorm(\lambda_{t-1}, \pi_t)$\\
    $p_{\lambda'} \gets P(\pi_{t}, \lambda'|X)$ \\
    $\lambda_t \gets \lambda'$ with probability $\beta=min(1,\frac{p_{\lambda_{t-1}}}{p_{\lambda'}})$}
\end{enumerate*}
\label{algo:bayesianMallows}
}
\end{algorithm}

\section{Supplementary results}\label{apd:results}

\subsection{Synthetic data}

We decided to introduce noise to the synthetic data to mimic what was seen in the healthcare data (see section \ref{details_PPA}). To simulate a missing percent of $q$ we randomly sampled a subset of $q$ individuals in the dataset. We then artificially truncated their ranking to reflect right censoring. We did this by selecting a rank position, $r_q$ in a normal range around three quarters of the way through the total ranking:
\begin{align*}
    r_q \sim Normal(0.75m, 1),
\end{align*}
where $m$ is the length of the ranking. Then for all ranks in a position greater than or equal to $r_q$ we deleted the rank information.


\subsection{Healthcare data}\label{details_PPA}

Table \ref{table:dataset} lists the subset of symptoms we included in the model. Of the $n=30$ respondents $27$ were caregivers, or bereaved caregivers, to people living with PPA, and three were responses recorded by people living with PPA. 

When answering the PPA questionnaire, respondents were told not to respond to questionnaires relating to symptoms that they had yet to experience. As a result there was considerable right censoring of the data, as most respondents were at a middle disease stage, and thus yet to experience symptoms associated with the latter disease course.

\begin{table}[h]
    \small
    \centering
    \begin{tabular}{p{4.5cm}p{1cm}p{1cm}}
        \toprule
        Symptom & Prior order & \% responded \\
        \midrule
        \midrule
        Changes to sleeping patterns, e.g. napping & 1  & 76.7 \\
        \midrule
        Gluttonous & 2 & 63.3\\
        \midrule
        Bodily complaints with no apparent cause & 2 & 56.7\\
        \midrule
        Increased sensitivity to sound / tinnitus & 2 & 43.3 \\
        \midrule
        More `rigid' / obsessional & 2 & 60.0 \\
        \midrule
        Walking more slowly & 3 & 53.3 \\
        \midrule
        Needs help dressing & 3 & 60.0 \\
        \midrule
        Difficulty swallowing & 4 & 36.7 \\
        \bottomrule
    \end{tabular}
\caption{The list of well-being symptoms we used in the model. We had responses for $n=30$ individuals with svPPA, collected from caregivers (and bereaved caregivers).}
\label{table:dataset}
\end{table}

\end{document}